\documentclass{article}

    \usepackage[final]{neurips_2022_ml4ad}

\usepackage[utf8]{inputenc} 
\usepackage[T1]{fontenc}    
\usepackage{hyperref}       
\usepackage{url}            
\usepackage{booktabs}       
\usepackage{amsfonts}       
\usepackage{nicefrac}       
\usepackage{microtype}      
\usepackage{xcolor}         
\usepackage{graphicx}
\usepackage{subfigure}
\usepackage[ruled,vlined]{algorithm2e}
\usepackage{adjustbox}
\usepackage{multirow}
\usepackage{amsmath}
\usepackage{bm}

\title{Potential Energy based Mixture Model for Noisy Label Learning}

\author{%
  Zijia Wang\\
  Dell Technologies\\
  Shanghai, China\\
  \texttt{Zijia\_Wang@Dell.com} \\
  \And
  Wenbin Yang \\
  Dell Technologies\\
  Shanghai, China\\
  \texttt{ralph.yang@dell.com} \\
  \And
  Zhisong Liu\\
  Dell Technologies\\
  Shanghai, China\\
  \texttt{zhisong.liu@dell.com} \\
  \And
  Zhen Jia\\
  Dell Technologies\\
  Shanghai, China\\
  \texttt{z\_jia@dell.com} \\
}

\begin{document}

\maketitle

\begin{abstract}
Training deep neural networks (DNNs) from noisy labels is an important and challenging task. However, 
most existing approaches focus on the corrupted labels and ignore the importance of inherent data structure. To bridge the gap between noisy labels and data, inspired by the concept of potential energy in physics, we propose a novel Potential Energy based Mixture Model (PEMM) for noise-labels learning. We innovate a distance-based classifier with the potential energy regularization on its class centers.
Embedding our proposed classifier with existing deep learning backbones, we can have robust networks with better feature representations. They can preserve intrinsic structures from the data, resulting in a superior noisy tolerance.
We conducted extensive experiments to analyze the efficiency of our proposed model on several real-world datasets. Quantitative results show that it can achieve state-of-the-art performance. 
\end{abstract}

\section{Introduction}

With the recent emergence of large-scale datasets, deep neural networks (DNNs) have exhibited impressive performance in numerous machine learning tasks\cite{wang2019symmetric}. However, labeling large-scale datasets is a costly and error-prone process, and even high-quality datasets are likely to contain noisy labels \cite{2018Iterative}. 
Recent studies~\cite{arazo2019unsupervised,wang2019symmetric} show that noisy labels severely degrade the generalization performance of deep neural networks.
Therefore, learning from noisy labels 
, or we also refer it as robust learning, 
is becoming an important task. 

Currently, several works focus on 
robust learning mechanisms to reduce the effects caused by noise.
There are many approaches that resolve the problem from different aspects. For example, \cite{zhang2018generalized,wang2019symmetric,ghosh2017robust} propose noise-robust losses and robust regularizers to avoid exceptions caused by outliers. \cite{jiang2018mentornet,2017Decoupling,2018Iterative} realize co-teaching deep neural networks for knowledge transfer. \cite{2015Learning,vahdat2017toward,li2017learning} utilize label correction to improve the data robustness.
However, these approaches only 
focus on the technologies that separate the clean data from the noisy ones, and ignore the inherent information of data samples, which may cause over-confidence 
on noisy labels\cite{wang2019symmetric,li2020dividemix}.

In this paper, we investigate the problem in a ``natural way". Our method is motivated by the following
observation: clean datasets are often low-rank\cite{li2019gradient}, which means they are in a ``stable state" with the lowest entropy \cite{yu2020learning}, while the existence of noisy label will break the inherent low-rank state. 
Coincidentally, things in the world always tend to exist in such a steady state because of the potential energy (PE) among molecules.
Similar phenomenon also exists in the real world where the potential energy (PE) keeps the molecules in a steady state.

\begin{figure}[ht]
\centering
\includegraphics[width=0.50\columnwidth]{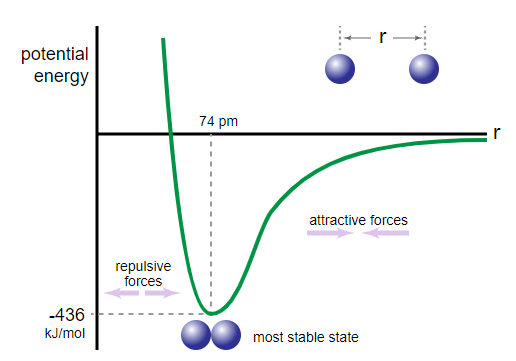} 
\caption{An illustration of potential energy.  The energy of the molecule is a function of the distance between the two nuclei. The most stable state(potential energy is -436KJ/mol) is achieved at the distance 74 pm.}
\label{fig:PE}
\end{figure}

The mechanism of PE is illustrated in Fig. \ref{fig:PE}: for two nuclei, there exists a distance that makes the system most stable, while in this state, the potential energy becomes minimal. Inspired by this theory, we propose a Potential Energy based Mixture Model (PEMM) for robust learning. 
Concretely, we propose a distance based classifiers constrained by a novel Potential Energy based Mixture Model (PEMM). It could preserve the inherent structures of data samples, so the training process becomes less dependent on class labels. In the Experiment section, we will demonstrate the strengths of our approach on several datasets. We will also provide ablation studies for detailed analysis.

In summary, the advantages and contributions of the algorithm proposed in this paper are:
\begin{itemize}
    \item \textbf{An explainable noisy labels learning method.} Compared with other methods with hand-crafted combined loss, our PEMM framework is based on essence of the machine learning, which makes the algorithm more explainable. 
    \item \textbf{Out of box.} Compared to existing approaches that often involve architectural or non-trivial algorithmic modifications, our method is easy to use. The PEMM can be easily incorporated into existing methods to significantly improve their performance.
    \item \textbf{Satisfied performance.} Empirically, it is demonstrated that the PEMM can achieve better robustness than state-of-the-art methods. 
\end{itemize}

\section{Related Work}
\label{sec::overview}
\subsection{Noisy-robust training}
Several methods that aim to mitigate label noise have been proposed. Here, we summarize some of the recent approaches to noise-robust learning. For simply, this method can be roughly divided into three categories: 1) label correction , 2) loss adjustment , and 3)  training strategies modification.

\textbf{Label correction} One common label correction approach is to  correct data labels via a clean label inference step using complex noise models characterized by some probabilistic models, like graphical models \cite{2015Learning}, conditional random fields \cite{vahdat2017toward} or knowledge graphs \cite{li2017learning}. However, the ground-truth noise models are neither always available in practice nor easy to obtain accurate estimation. 

\textbf{Loss adjustment} A family of studies modify the regularization  to improve the robustness.  For example,  PHuber\cite{menon2019can}  proposes a composite loss-based gradient clipping for label noise robustness. Moreover, Label Smoothing Regularization (LSR)\cite{szegedy2016rethinking} adopts soft labels to avoid overfitting to noisy labels. Recently, the mixup\cite{zhang2017mixup} regularizes the DNN to favor simple linear behaviors in raining examples.

Meanwhile, researchers have attempted to design robust loss functions. Work in \cite{ghosh2017robust} shows that Mean Absolute Error (MAE) is theoretically robust against label noise. However, MAE may lead to significantly longer convergence time and performance drop\cite{zhang2018generalized}. To overcome the problem, Generalized Cross Entropy (GCE)\cite{zhang2018generalized} loss applies a Box-Cox transformation to probabilities can behave like a weighted MAE. Furthermore, Wang  proposed Symmetric Cross Entropy loss (SCE)\cite{wang2019symmetric}, which combines CE loss with Reverse Cross Entropy loss. 


\textbf{Training strategies modification} There are some methods design new learning paradigms for noisy labels. Work in \cite{jiang2018mentornet} proposed a framework that supervises the training of a student-net by a learned sample weighting scheme in favor of probably correct labels. Furthermore, the iterative learning framework \cite{2018Iterative} iteratively detects and isolates noisy samples during the learning process. However, these methods either rely on complex interventions into the learning process, which may be challenging to adapt and tune.

\subsection{Larger margin learning}
The concept of large margin learning aims to keep the value of empirical risk fixed and minimize the confidence interval\cite{Vapnik1999An}. The classification margin is closely related to adversarial robustness and generalization ability.
However, larger margin learning methods, like large margin cosine loss(LMCL)\cite{wang2018cosface} and additive angular margin loss(AAML)\cite{2018ArcFace}, are not developed for noisy labels  problems directly and  their decision boundaries are affected by noisy labels. 

\section{Problem Statement}
\label{sec:problem}

Given a $K$-class dataset $\mathcal{D} = \left\{\bm{x}; y\right\}^N_{n=1}$, with $\bm{x} \in \bm{X} \subseteq R^d$ 
and $y\in  \mathcal{Y} = \left\{1;...; K\right\} $ representing a $d$-dimensional input and its associated label respectively. For each sample $\bm{x}$, a classifier $f(\bm{x})$ output its probability of each label $k  \in \left\{1;... ; K\right\}: p(k|\bm{x})$. Denoting the ground-truth distribution over labels for sample $\bm{x}$ by $q(k|\bm{x})$, and $\sum_{k=1}^K q(k|\bm{x}) = 1$. Then the cross entropy (CE) loss for the data sample $\bm{x}_n$ can be expressed as:

\begin{equation}
\label{eq:ce}
    l_{ce}(\bm{x}_n) = -\sum_{k=1}^{K} q(k|\bm{x}_n) \log(p(k|\bm{x}_n))
\end{equation}

Although easy to converge, the CE loss may easily be overfit to noise labels \cite{wang2019symmetric}. In CE, samples with less congruence on provided labels are implicitly weighed more in the gradient update. This implicit weighting scheme is desirable for training with clean data, but can cause overfitting to noisy labels. 


\section{Solution}
The main idea behind the proposed method is that the inherent data structure can be modeled by fitting a mixture model. Then, with representations that preserve intrinsic structures from the data, training relies less on class labels and becomes more robust.  


\subsection{Inherent data information}
In general, the machine learning (ML) tasks are trying to estimate the real data distribution $p(\bm{x})$ with a parametric model. Formally, we rewrite the estimation process with latent feature $\bm{z}$ and data label $y$ as
\begin{equation}
\label{eq:full}
\begin{aligned}
    & KL\big(p(\bm{x},\bm{z},y)\big\Vert q(\bm{x},\bm{z},y)\big) = \sum_y p(\bm{z},y|\bm{x})\tilde{p}(\bm{x})\ln \frac{p(\bm{z},y|\bm{x})\tilde{p}(\bm{x})}{q(\bm{z},y|\bm{x})\tilde{q}(\bm{x})} d\bm{z}d\bm{x}
\end{aligned}
\end{equation}
where $\tilde{p}(\bm{x})$ and $\tilde{q}(\bm{x})$ are the data distribution estimated with finite data samples and the given data distribution. From our problem, we can set $\tilde{p}(\bm{x})= \tilde{q}(\bm{x})$.

Then, in a deep learning framework, it is natural to assume that
\begin{equation}
\begin{aligned}
        p(\bm{z},y|\bm{x})& =p(y|\bm{z})p(\bm{z}|\bm{x}),\quad q(\bm{z},y|\bm{x})& =q(\bm{z}|y)q(y|\bm{x})
\end{aligned}
\end{equation}
Then, we can rewrite the eq.\ref{eq:full} as a form of expectation:
\begin{equation}
\label{eq:exp1}
\begin{aligned}
     \mathbb{E}_{\bm{x}\sim\tilde{p}(\bm{x})}\Big[\sum_y p(y|\bm{z}) \log& \frac{p(y|\bm{z})p(\bm{z}|\bm{x})}{q(\bm{z}|y)q(y|\bm{x})} \Big],  \quad \bm{z}\sim p(\bm{z}|\bm{x}) \\
\end{aligned}
\end{equation}
where $\bm{z}\sim p(\bm{z}|\bm{x})$ is the reparameterize operation. One can further rewrite the eq. \ref{eq:exp1} as

\begin{equation}
\label{eq:exp2}
\begin{aligned}
    \mathbb{E}_{\bm{x}\sim\tilde{p}(\bm{x})} &\Big[ -\sum_y p(y|\bm{z}) \log q(\bm{z}|y)- \sum_y p(y|\bm{z})\log q(y|\bm{x})\\
    & + \sum_y p(y|\bm{z})\log p(z|\bm{x}) + \sum_y p(y|\bm{z})\log p(y|\bm{z}) \Big],  
     \bm{z}\sim p(\bm{z}|\bm{x})
\end{aligned}
\end{equation}

where $q(y|\bm{z})\log p(\bm{z}|y)$ is essentially the CE loss, $p(y|\bm{z})\log q(y|\bm{x})$ is the loss when given a reference distribution $p(y|\bm{x})$, $p(y|\bm{z})\log p(z|\bm{x})$ will be minimized when the features are well separated, $p(y|\bm{z})\log p(y|\bm{z})$ is a constant.
    
From eq.\ref{eq:exp2}, one can find that the widely used CE loss only uses part of the information. In the context of noisy labels, $q(y|x)$ does not represent the true class distribution well. 

\begin{figure}[htbp]
    \centering
    \includegraphics[width=0.7\columnwidth]{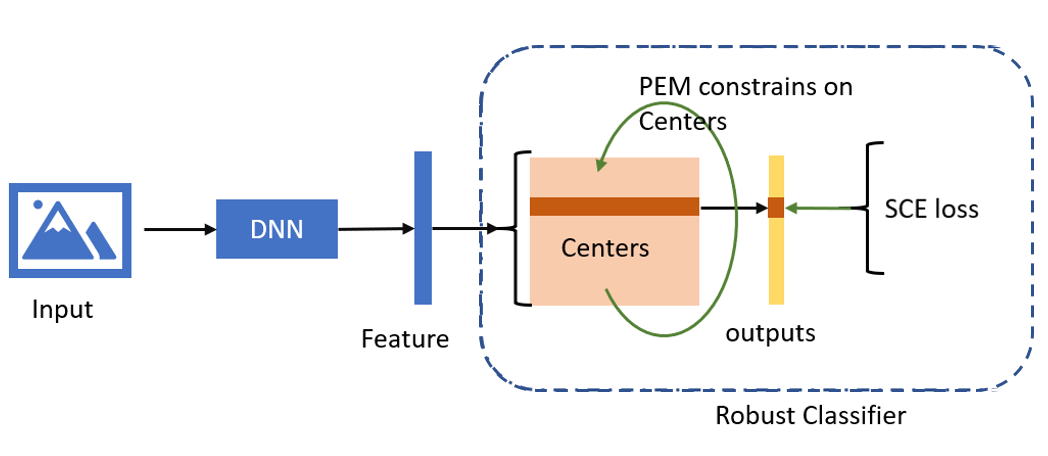}
    \caption{The framework for noisy robust learning.}
    \label{fig:frame}
\end{figure}

To fully utilize the inherent structures information of the data, i.e. approximate the expectation in eq.\ref{eq:exp2},  we propose the learning framework shown in Fig. \ref{fig:frame}. The classifier(section \ref{sec::clf}) with  Potential Energy (PE) regularizer (section \ref{sec:PEMcenter}) on its class centers is utilized for noisy robust learning.

\subsection{Distance based classifier}
\label{sec::clf}
In some recent works about few-shot learning \cite{fort2017gaussian}, it is reported that the distance based classifier is more explainable. Assuming the $\bm{C}=[\bm{c}_1,\dots,\bm{c}_K]^T$ is the matrix containing all the class centers, the distance between sample $\bm{x}_{n}$ and the $k$-th center vector $\bm{c}_{k}$ can be expressed as:
\begin{equation}
    d_{n,k} = exp(- \frac{\left\Vert \bm{z}_n - \bm{c}_k \right\Vert_2^2}{\sigma^2})
\label{eq:dd}
\end{equation}
where $\bm{z}_n=f(\bm{x}_n)$ is the latent space feature before classifier, $\|  \|_2$ stands for the $l_2$ norm of a vector, $\bm{c}_k$ is the center of  $k$-th class, and $\sigma$ is the hyper-parameter of the radius of the Euclidean ball. In the sequence, the posterior probability of $\bm{x}_n$ can then be calculated as

\begin{equation}
    p(y_{n,k}|\bm{x}_n) = \frac{d_{n,k}}{\sum_{k=1}^{K} d_{n,k}}
\end{equation}


Then, we noticed that, the second term in eq.\ref{eq:exp2} reflect the penalty of distribution $q(y|\bm{x})$ given the prediction distribution $p(y|\bm{z})$. Similar to \cite{wang2019symmetric}, one can use Reverse Cross Entropy (RCE) to measure the penalty effectively. The RCE loss for a sample $\bm{x}$ is:
\begin{equation}
\begin{aligned}
        l_{rce}(\bm{x}_n)  = - \sum_{k=1}^{K} p(k|\bm{x}_n) \log(q(k|\bm{x}_n))
\end{aligned}
\label{eq:rce}
\end{equation}
In the sequence, we adopt the RCE loss\cite{wang2019symmetric} and give the final classifier loss as
\begin{equation}
\begin{aligned}
        l_{clf}(\bm{x}_n)  = & -\alpha \sum_{k=1}^{K} q(k|\bm{x}_n)\log(p(k|\bm{x}_n)) - \sum_{k=1}^{K} p(k|\bm{x}_n) \log(q(k|\bm{x}_n))
\end{aligned}
\label{eq:class}
\end{equation}
where $\alpha$ is the parameter for flexible exploration on the robustness. Intuitively, the $l_{clf}$ will cover the first term and second term in eq. \ref{eq:exp2}. 


\subsection{Potential energy based centers regularization}
\label{sec:PEMcenter}

To cover the third term in eq.\ref{eq:exp2}, we introduce the potential energy based center regularization. Potential energy is a concept in physics. As shown in Fig. \ref{fig:PE}, there exists a distance($r_0$) that makes the system most stable \cite{mccall2010physics}. We adapt this concept and establish the stable state for robust feature learning \cite{li2017learning,yu2020learning}. 


In this paper, we use the potential energy expression:
\begin{equation}
    E(r) = \frac{1}{r^{u}} - \frac{\xi}{r^{v}}
\label{eq:er}
\end{equation}
where $r>0$ is the distance between any two nuclei, $v \in \mathbb{N}^+$ and $u \in \mathbb{N}^+$ represents the power of push and attractive force, respectively. $\xi$ is the weight of attractive force. For computing effectiveness, we choose $u=3$ ,$v=2$ and $\xi=2$ for a approximation to the real-word potential energy curve \cite{2009textbook}. The $r_0$ under these settings is calculated as 0.75.

Then we adapt this equation into the loss function of class centers matrix $\bm{C}=[\bm{c}_1;...;\bm{c}_K]$ as :

\begin{equation}
\begin{aligned}
 {L}_{pem}(\bm{C}) = & \sum_{i=0}^{K-1} \sum_{j=i+1}^{K} \big[{l}_{pem}(\bm{c}_i,\bm{c}_j)\big]
\end{aligned}
\label{eq:pem}
\end{equation}

where
\begin{equation}
\begin{aligned}
 &{l}_{pem}(\bm{c}_i,\bm{c}_j) =\frac{1}{( \|\bm{c}_i-\bm{c}_j\|_2+\beta)^{3}}- \frac{2}{( \|\bm{c}_i-\bm{c}_j\|_2 + \beta) ^{2}}
\end{aligned}
\label{eq:dis}
\end{equation}
where $\bm{c}_0 = \bm{0}$; $K>2$ is the number of classes to be compared, $0<\beta<r_0$ are the hyper-parameters to control the loss value.

\begin{figure}[h]
    \centering
    \includegraphics[width=0.7\columnwidth]{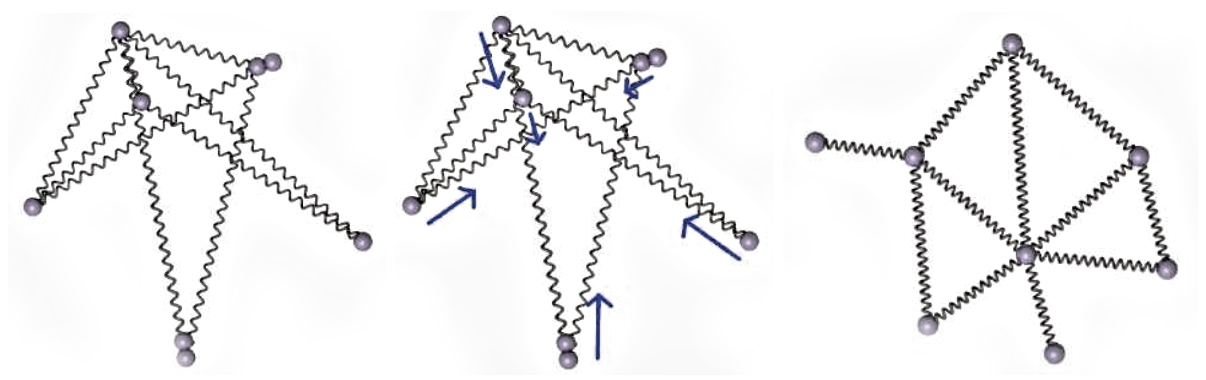}
    \caption{The establishment process of co-stable state: starting from random positions, then each node will look for a stable position automatically.}
    \label{fig::PEprocess1}
\end{figure}

Fig. \ref{fig::PEprocess1} illustrates the effectiveness of PE constrains. The PE constrains try to make the system stable by minimizing the potential energy. After some iterations, the system will achieve “aesthetically pleasing” layout, which can be captured by two criteria: "all the edge lengths ought to be the same, and the layout should display as much symmetry as possible." \cite{2012Spring}.

\subsection{Training}
\label{sec::train}
The loss function covers all terms in eq. \ref{eq:exp2}, then it can be expressed as:

\begin{equation}
\label{eq:loss}
    L = \sum_{n=1}^B(t) \bigg[ L_{clf}(\bm{x}_n)\bigg] + \lambda L_{pem}(\bm{C})
\end{equation}
 where $B$ is a batch-size of samples, $\lambda$ is trade-off parameter to balance the contributions of the $l_{pem}$. 
Formally, our method can is express as Algorithm \ref{alg1}. It should be noticed that, one can also reweight the data samples according to the estimated distribution of our PEMM as the state-of-the-arts\cite{li2020dividemix,yao2020searching}.

\begin{algorithm}[!htbp]
\label{Alg1}

\SetKwInOut{KIN}{Input}
\SetKwInOut{KOUT}{Output}
\caption{Robust training of PEMM}
\KIN{Noisy dataset $\{{\bm{x},y}\}\in \mathcal{D}$;epoch number ${T}$;learning rate $\delta$ }
\KOUT{Robust model}

Initialize: Network parameters  $\bm{\Theta}$

\For{$t \leftarrow 1$ to $T$} {
    \For{each mini batch B}{
    obtain features $\bm{z}$\\ 
    compute distance ${\bm{d}}$ (eq.\ref{eq:dd})\\
    compute the classify loss ${l_{clf}}$ (eq.\ref{eq:class})\\
    compute potential energy loss ${l_{pem}}$(eq.\ref{eq:pem})  \\
    compute  loss ${L}$ of the batch (eq.\ref{eq:loss})  \\
    update $\bm{\Theta}$ using back propagation\\
    
    }
}
\textbf{return}

\label{alg1}
\end{algorithm}

\section{Comparison with large margin learning }
\label{sec::Anna}
\begin{figure}[h]
    \centering
    \includegraphics[width=0.35\columnwidth]{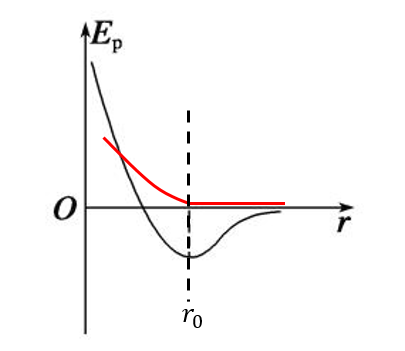}
    \caption{Illustration of the difference between the loss in \cite{chen2019joint}(red) and our PE loss(black). The PE loss value is asymmetric.}
    \label{fig::EN}
\end{figure}

Intuitively, PE based class center regularization is similar to a larger margin learning methods\cite{guo2021recent}. The larger margin learning will enforce large margins among centers across different categories. 
However, different form  the well-known large-margin loss \cite{guo2021recent}, which utilize the pre-defined boundary, our method is more flexible by directly minimize the potential energy.
Moreover, a class center based discriminative loss based on Euclidean distance is also proposed in \cite{chen2019joint}:
\begin{equation}
\label{eq:oc}
   l_{sq}(\bm{c}_i,\bm{c}_j)=max(0,r_0-\|\bm{c}_i-\bm{c}_j\|_2 ).
\end{equation}
As shown in the Fig.\ref{fig::EN}, the loss proposed in \cite{chen2019joint} and eq.\ref{eq:dis} are both related to the distance $r_0$. However, for the loss in eq.\ref{eq:oc}, when the class distance bigger than $r_0$, the penalty  will no longer change, while the PE loss is asymmetric and only reaches its minimization (most stable state) at $r_0$.

\begin{figure}[h]
    \centering
    \includegraphics[width=0.6\columnwidth]{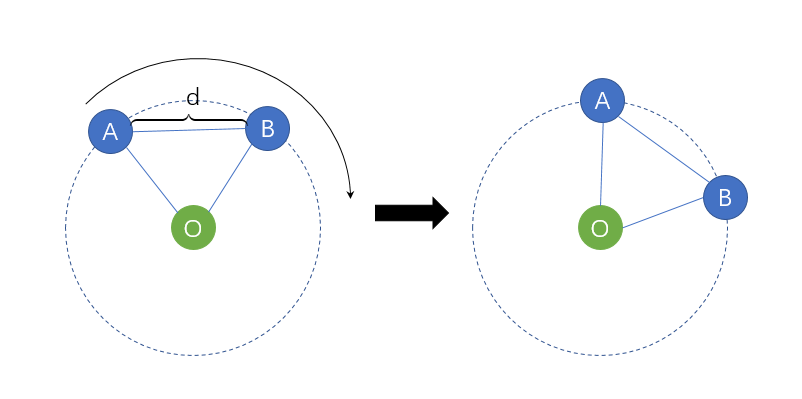}
    \caption{Change process of a co-stable process.(left to right)}
    \label{fig::Change}
\end{figure}

The asymmetric property of PE is important to establish a co-stable system. In the high-dimensional space, the PE loss will help the class center to form such a polyhedron whose distance between any two vertices keep the same. Obviously, such a polyhedron system is co-table. Without loss of generality, consider a simple binary classification system shown in the Fig.\ref{fig::Change}. Due to the PE loss, the distance between A, B, and O (coordinate origin) are $d=r_0-\beta$. When change the position of class A, 
one has to change the position of class B at the same time. Obviously, 
these changes will be equivalent to rotating A and B at the same time. So, such a co-stable state system makes full use of complementary information between classes \cite{song2021learning}. In addition, such a co-stable system is also an output of implicit early-stopping mechanism. According to eq.\ref{eq:pem}, the limitation of the sum of class centers distances as
\begin{equation}
\label{eq:limit}
   \sum_{i=0}^{K-1} \sum_{j=i+1}^K \|\mathbf{c}_i - \mathbf{c}_j\|_2 \Rightarrow \frac{K(K+1)}{2} (r_0-\beta).
\end{equation}

The memorization nature \cite{li2019gradient} indicates that the DNNs can avoid overfitting to noisy label by early stopping. 
For our model, the PE naturally result in an equivalent early stooping due to the limitation in eq.\ref{eq:limit}, which will make the training process robust to the noisy labels.

\section{Experiment}
\label{sec::exp}
In this section, we describe the experiments performed to evaluate our method. We verify our method by comparing with other recent baseline methods, varying experimental settings in terms of dataset and type and ratio of noise in the training data.

\subsection{Empirical Understanding}
We conduct experiments on CIFAR-10 dataset with symmetric noise towards a deeper understanding of PEMM. Symmetric noisy labels are generated by flipping the labels of a given proportion of training samples to one of the other class labels uniformly. 

\textbf{Experimental setup:} An 8-layer DNN with 6 convolutional layers followed by 2 fully connected layers are tested in our experiment. All networks are trained using SGD with momentum 0.9, weight decay 10$^{-4}$. The initial learning rate is established as 0.01, which is divided by 10 after 40 and 80 epochs (120 epochs in total). In our network setting , all the features $\bm{z}$ will be normalized before eq.\ref{eq:dd}, and we will set $\sigma$ to 1 in all experiments. The parameter $\alpha$, $\beta$ and $\lambda$ in PEMM are set to 0.1,0.3,1 respectively.

\begin{figure}[h]
\centering
\subfigure[]{
\includegraphics[width=0.3\columnwidth]{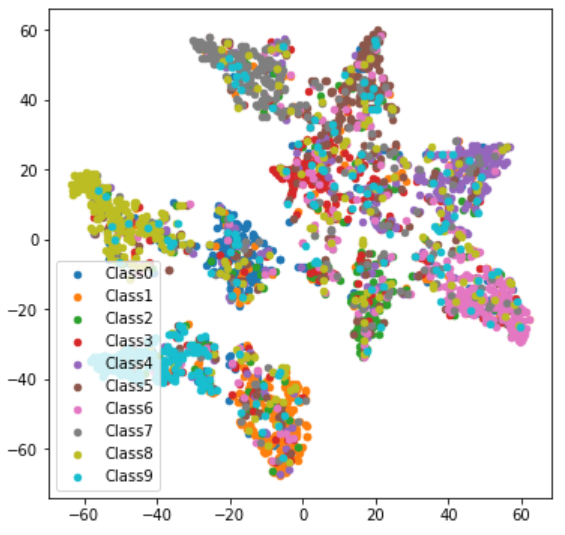}
}
\quad
\subfigure[]{
\includegraphics[width=0.3\columnwidth]{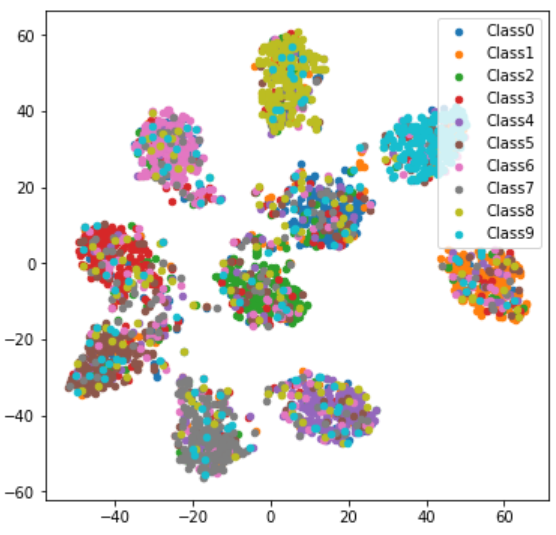}
}
\quad
\subfigure[]{
\includegraphics[width=0.3\columnwidth]{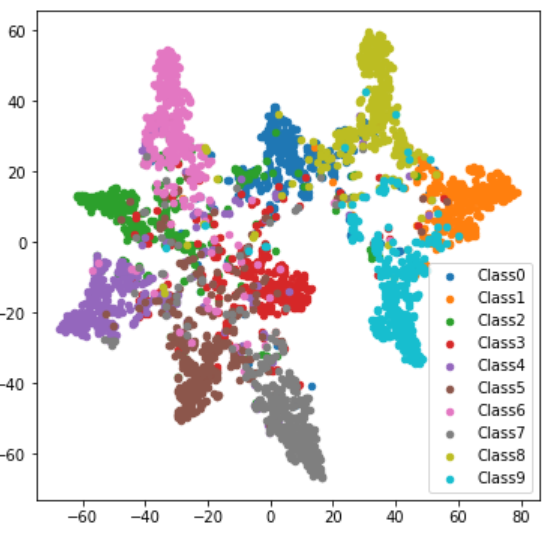}
}
\quad
\subfigure[]{
\includegraphics[width=0.3\columnwidth]{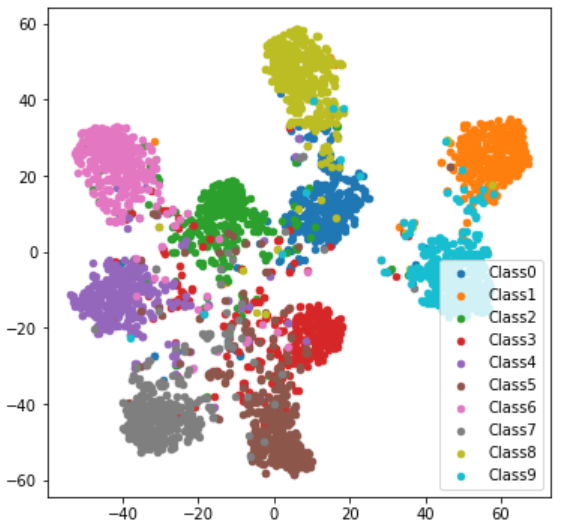}
}
\caption{Representations learned by CE and PEMM with 40\% symmetric noisy labels for training ,(top) and validation (bottom) in CIFAR-10 dataset with (a)(c) cross-entropy loss from Eq. 1 (test accuracy 78.7\%), (b)(d) the proposed PEMM (test accuracy 87.9\%).}
\label{fig::feavis}
\end{figure}

\textbf{Representations:} We investigate the representations learned by PEMM compared to that learned by CE. The t-SNE\cite{van2008visualizing} are utilized to project  the representation at the second last dense layer to a 2D. The projected representations for 40\% noisy labels are shown in Fig. \ref{fig::feavis} .One can easily found that,  the representations learned by PEMM have significantly better quality than that of CE with more separated class centers and clearly bounded clusters.

\begin{figure}[h]
\centering
\subfigure[]{
\includegraphics[width=0.3\columnwidth]{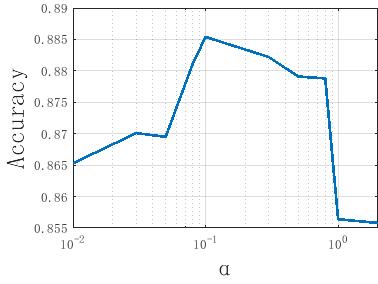}
}
\quad
\subfigure[]{
\includegraphics[width=0.3\columnwidth]{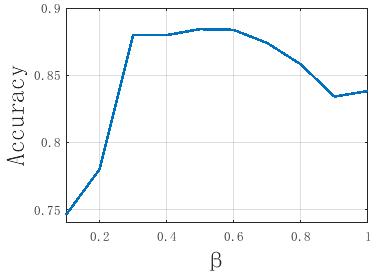}
}
\quad
\caption{The effect of different values of $\alpha$(a) and $\beta$(b).}
\label{fig::pa}
\end{figure}

\textbf{Parameter analysis:} We tune the parameters of PEMM: $\alpha$, $\beta$ and $\lambda$. As $\lambda$ can be reflected by $\beta$, thus we only show results of $\alpha$ and $\beta$. We tested $\alpha \in [0.01; 1]$ and $\beta \in [0.1; 1]$ on CIFAR-10 under 40\% noisy labels. Fig.\ref{fig::pa} (a) shows that large $\alpha$ (e.g., 1.0) tends to cause
more overfitting, while small $\alpha$ (e.g., 0.01) will some the robustness to noisy. For this reason, a relatively large $\alpha$ may help to convergence on more complex datasets.
As for parameter $\beta$, one can see that PEMM is not very sensitive to $\beta$ (Fig.\ref{fig::pa} (b)) when $0.2<\beta<r_0$. However, if $\beta<0.2$ or $\beta>r_0$, the performance of our PEMM will decrease due of the properties of PE function.

\begin{figure}[htbp]

\centering
\subfigure[Remove RCE loss loss (accuracy 83.72\%)]{
\includegraphics[width=0.28\columnwidth]{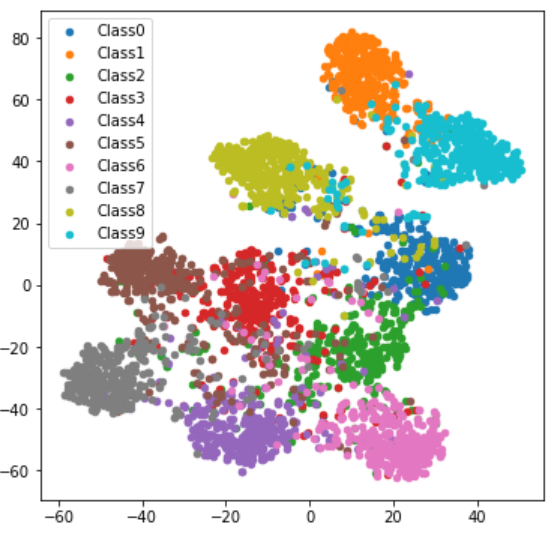}
}
\quad
\subfigure[Remove CE loss (accuracy 86.50\%)]{
\includegraphics[width=0.28\columnwidth]{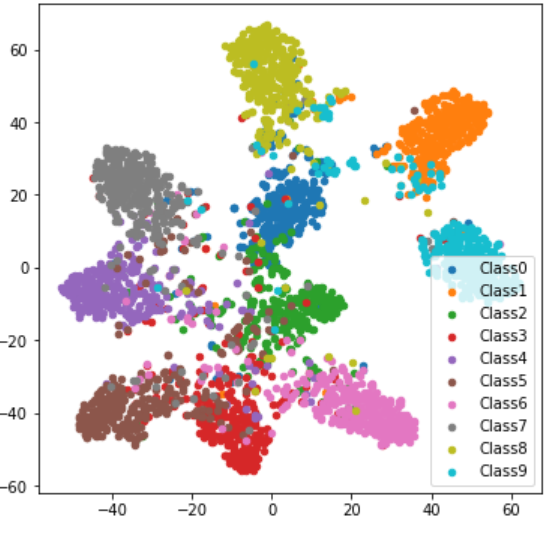}
}
\quad
\subfigure[Remove the PEM term (accuracy 83.80\%)]{
\includegraphics[width=0.28\columnwidth]{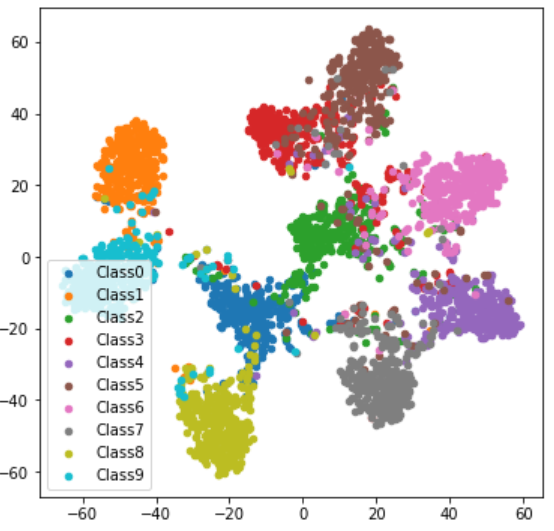}
}

\caption{Ablation experiments with 40\% symmetric noisy labels for  in CIFAR-10 train dataset.All the t-SNE visualization in performed in test dataset}
\label{fig::study}
\end{figure}

\textbf{Ablation study:} For a comprehensive understanding of each component in PEMM, we further conduct a series of ablation experiments on CIFAR-10 under 40\% noisy labels.  Figure\ref{fig::study} presents the following experiments: a) replace SCE with CE loss ; b) remove CE loss; c) remove the PEM term; d) remove the reconstruction term. 

For a better understanding, we visualize the feature distribution with t-SNE again. In Fig \ref{fig::study}(a), one can observe that even without SCE loss function, our method still has better noisy robustness than the CE-loss based method. Then, compare Fig.\ref{fig::study}(a) and Fig.\ref{fig::study}(b), the reverse-CE term in SCE loss has more impact on the performance improvement. Moreover, it is witnessed in \ref{fig::study}(c) that the PEM loss term can enlarge the decision boundary. 

\subsection{Robustness to Noisy Labels}

We compare PEMM with 4 recently proposed noisy label learning methods as well as the standard CE loss: (1) Forward \cite{patrini2017making}: Training with label correction by multiplying the network prediction with the ground truth noise matrix;(2) Bootstrap \cite{reed2014training}: Training with new labels generated by a convex combination of the raw labels and the predicted labels; (3) GCE \cite{zhang2018generalized}: Training with a noise robust loss encompassing both MAE and CE;  (4) SCE \cite{wang2019symmetric}: Training with the novel symmetric cross entropy loss;and (5) CE: Training with standard cross entropy loss.

\begin{table*}[htbp]

\caption{Test accuracy (\%) of different models on benchmark datasets with various rates of symmetric and asymmetric noisy labels. The average accuracy and standard deviation of 5 random runs are reported and the best results are in \textbf{bold}.}

\label{tab:all_acc}
\centering
\small
\resizebox{\textwidth}{!}{
\begin{tabular}{c|c|ccccc|ccc}
\hline
\multirow{3}{*}{Datasets} & \multirow{3}{*}{Methods} & \multicolumn{5}{c}{Symmetric Noise} & \multicolumn{3}{c}{Asymmetric Noise} \\ \cline{3-10}
 &  & \multicolumn{5}{c}{Noise Rate } & \multicolumn{3}{c}{Noise Rate } \\
& & 0.0 & 0.2 & 0.4 & 0.6 & 0.8 & 0.2 & 0.3 & 0.4 \\ \hline \hline

\multirow{7}{*}{CIFAR-10} & CE & 89.26 $\pm$ 0.03 & 82.96 $\pm$ 0.05 & 78.70 $\pm$ 0.07 & 66.62 $\pm$ 0.15 & 34.80 $\pm$ 0.25 & 85.98 $\pm$ 0.03 & 83.53 $\pm$ 0.08 & 78.51 $\pm$ 0.05 \\
 & Bootstrap & 88.77 $\pm$ 0.06 & 83.95 $\pm$ 0.10 & 79.97 $\pm$ 0.07 & 71.65 $\pm$ 0.05 & 41.44 $\pm$ 0.49 & 86.57 $\pm$ 0.08 & 84.86 $\pm$ 0.05 & 79.76 $\pm$ 0.07 \\
 & Forward & 89.39 $\pm$ 0.04 & 85.83 $\pm$ 0.05 & 81.37 $\pm$ 0.03 & 73.59 $\pm$ 0.08 & 47.10 $\pm$ 0.14 & 87.68 $\pm$ 0.01 & $\boldsymbol{86.86 \pm 0.06}$ & ${85.73 \pm 0.04}$ \\
 & GCE & 86.76 $\pm$ 0.03 & 84.86 $\pm$ 0.06 & 82.42 $\pm$ 0.10 & 75.20 $\pm$ 0.09 & 40.81 $\pm$ 0.24 & 84.61 $\pm$ 0.09 & 82.11 $\pm$ 0.13 & 75.32 $\pm$ 0.10 \\
  & SCE & $\boldsymbol{91.05 \pm 0.04}$ & $\boldsymbol{90.88 \pm 0.06}$ & $\boldsymbol{86.34 \pm 0.07}$ & $\boldsymbol{80.07 \pm 0.02}$ & $\boldsymbol{53.81 \pm 0.27}$ & $\boldsymbol{88.24 \pm 0.05}$ & 85.36 $\pm$ 0.14 & 80.64 $\pm$ 0.10 \\
 & \textbf{PEMM} & $\boldsymbol{89.28 \pm 0.04}$ & $\boldsymbol{89.01 \pm 0.06}$ & $\boldsymbol{88.54 \pm 0.07}$ & $\boldsymbol{85.25 \pm 0.03}$ & $\boldsymbol{59.34 \pm 0.28}$ & $\boldsymbol{89.05 \pm 0.03}$ & 87.11 $\pm$ 0.15 & $\boldsymbol{85.78 \pm 0.15}$ \\ \hline
\hline
\multirow{7}{*}{CIFAR-100} & CE & 64.34 $\pm$ 0.37 & 59.26 $\pm$ 0.39 & 50.82 $\pm$ 0.19 & 25.39 $\pm$ 0.09 & 5.27 $\pm$ 0.06 & 62.97 $\pm$ 0.19 & 63.12 $\pm$ 0.16 & 61.85 $\pm$ 0.35 \\
 
 & Bootstrap & 63.26 $\pm$ 0.39 & 57.91 $\pm$ 0.42 & 48.17 $\pm$ 0.18 & 12.27 $\pm$ 0.11 & $1.00 \pm 0.01$ & $63.44 \pm 0.35$ & $63.18 \pm 0.35$ & $62.08 \pm 0.22$ \\
 & Forward & 63.99 $\pm$ 0.52 & 59.75 $\pm$ 0.34 & 53.13 $\pm$ 0.28 & 24.70 $\pm$ 0.26 & $2.65 \pm 0.03$ & $64.09 \pm 0.61$ & $64.00 \pm 0.32$ & $60.91 \pm 0.36$\\
 & GCE & $64.43 \pm 0.20$ & $59.06 \pm 0.27$ & $53.25 \pm 0.65$ & $36.16 \pm 0.74$ & $
 8.43 \pm 0.80$ & $63.03 \pm 0.22$ & $63.17 \pm 0.26$ & $61.69 \pm 1.15$ \\
 & SCE & $\boldsymbol{66.75 \pm 0.04}$ & $\boldsymbol{60.01 \pm 0.19}$ & $\boldsymbol{53.69 \pm 0.07}$ & $\boldsymbol{41.47 \pm 0.04}$ & $\boldsymbol{15.00 \pm 0.04}$ & $\boldsymbol{65.58 \pm 0.06}$ & $\boldsymbol{65.14 \pm 0.05}$ & $\boldsymbol{63.10 \pm 0.13}$\\
& \textbf{PEMM} & $\boldsymbol{67.00 \pm 0.05}$ & $\boldsymbol{62.10 \pm 0.18}$ & $\boldsymbol{65.75 \pm 0.12}$ & $\boldsymbol{56.25 \pm 0.05}$ & $\boldsymbol{30.03 \pm 0.05}$ & $\boldsymbol{65.78 \pm 0.08}$ & $\boldsymbol{65.80 \pm 0.03}$ & $\boldsymbol{64.81 \pm 0.18}$
\\ \hline
\end{tabular}}
\label{tab::result_table}
\vspace{-0.15 in}
\end{table*}

\textbf{Experimental setup:} We test two types of label noise: symmetric (uniform) noise and asymmetric (class-dependent) noise on CIFAR-10 \cite{krizhevsky2009learning} and CIFAR-100 \cite{krizhevsky2009learning} dataset. For asymmetric noisy labels, we flip labels within a specific set of classes.We flip the class labels as the setting in \cite{wang2019symmetric}.

We use the same network as Section 5.1 for CIFAR-10 , a ResNet-44 \cite{he2016deep} for CIFAR-100 and ResNet-32 \cite{he2016deep} for Imagenette. Parameters for the baselines are configured according to their original papers. For our PEMM, we have the same setting as Section 5.1. All networks are trained using SGD with momentum 0.9, weight decay 5 × 10$^{-3}$ and an initial learning rate of 0.1. The learning rate is divided by 10 after 20 and 50 epochs for CIFAR-10 (80 epochs in total),after 80 and 120 epochs for CIFAR-100 (150 epochs in total) and after 50 ,100 and 150 epochs for Imagenette (200 epochs in total). Simple data augmentation techniques (width/height shift and horizontal flip) are applied on all datasets. 

Next, we also test our proposed method on Clothing1M \cite{xiao2015learning}. The Clothing1M dataset contains 1 million images and the overall accuracy of the labels is ${\small\sim61.54\%}$ and may contain both symmetric and asymmetric label noise. The classification accuracy on the $10k$ clean testing data is used as the evaluation metric. 

The ResNet-50 with pretrained weights is used for Clothing1M dataset. We follow the same image processing setting in \cite{wang2019symmetric}. We train the models with batch size 64 and initial learning rate $10^{-3}$, which is reduced by $1/10$ after 5 epochs (10 epochs in total). SGD with a momentum 0.9 and weight decay $10^{-3}$ are adopted in our experiment. Other settings are the same as we did in CIFAR-100 dataset.

\textbf{Robustness performance:} All the results are shown in Table\ref{tab::result_table} and Table\ref{tab::nette}. 

In Table \ref{tab::result_table}, one can see that PEMM shows a clear improvement over other methods. This is likely because that,  other methods care too much about the noisy labels itself, while PEMM ensures richer representations that preserves intrinsic structures from the data, relies less on class labels.Moreover, the equivalent early stopping property also benefit our result and avoid over-learning on noisy class labels.

In Table \ref{tab::nette}, it is witnessed that our methods have competitive performance and does not require extra auxiliary information for noisy label correction.

\begin{table}[h]
\centering
\small
\caption{Accuracy (\%) of different models on real-world noisy dataset Clothing1M. The best results are in \textbf{bold}.}
\vspace{-0.1 in}
\label{tab::nette}
\begin{adjustbox}{width=0.5\columnwidth}
\begin{tabular}{l|ccccccc}
\hline
Methods & CE & Bootstrap & Forward &  GCE & SCE& PEMM\\ \hline
Acc & 68.80 & 68.94 & 69.84  & 69.75 & 71.02 & $\bm{72.80}$\\

\hline
\end{tabular}
\end{adjustbox}
\vspace{-0.15 in}
\end{table}

\section{Conclusions}
We present a PEMM based training algorithm that it can handle noisy labels without requiring priors from clean data.
A robust mixture model is proposed to approximate the latent distribution of data without supervision. It is implemented by the distance-based classifier with the PE loss on its class centers. 
We provide both theoretical and empirical analysis on PEMM, and demonstrate its effectiveness against various types and rates of label noise on both benchmark and real-world datasets. Overall, due to its simplicity and ease of implementation, we believe that PEMM 
shows great potential for noisy data training. In the future, it can be merged with other advanced techniques for real-world applications.
\bibliographystyle{apalike}
\bibliography{egbib}


\end{document}